%% file: main.tex
\documentclass[10pt,twocolumn,letterpaper]{article}

\usepackage[algorithms]{wacv}
\usepackage{algorithm}
\usepackage{algorithmic}


\usepackage{booktabs}
\usepackage{graphicx}
\usepackage[table]{xcolor}
\usepackage{multirow}

\definecolor{best}{RGB}{255,190,190}
\definecolor{second}{RGB}{255,242,170}
\definecolor{third}{RGB}{255,224,190}

\definecolor{wacvblue}{rgb}{0.21,0.49,0.74}
\usepackage[breaklinks,colorlinks,allcolors=wacvblue]{hyperref}

\def\wacvPaperID{624} 
\def\confName{WACV}
\def\confYear{2027}

\title{GC-GS: Optimizing 3D Gaussian Splatting with Gradient Conflict as a Geometry Signal}

\author{Chenhao Zhang$^{1}$ \qquad Fengze Zhang$^{2}$ \qquad Fengqing Zhu$^{1}$\\
$^{1}$Elmore Family School of Electrical and Computer Engineering, Purdue University\\
$^{2}$Department of Computer Science, Purdue University\\
West Lafayette, Indiana, USA
}

\begin{document}
\maketitle
\input{sec/0_abstract}    
\input{sec/1_intro}
\input{sec/2_related_work}
\input{sec/3_method}
\input{sec/4_experiment}

\input{sec/5_conclusion}
\clearpage
{
    \small
    \bibliographystyle{ieeenat_fullname}
    \bibliography{main}
}

\end{document}

%% file: sec/0_abstract.tex
\begin{abstract}
3D Gaussian Splatting (3DGS) is commonly optimized with multiple reconstruction losses, which naturally forms a multi-objective optimization problem. This issue becomes more pronounced when perceptual losses are introduced to improve visual quality, as different objectives may impose competing update directions on Gaussian primitives and make it difficult to simultaneously improve fidelity, perceptual quality, and representation efficiency. Based on this view, we reformulate 3DGS training as a multi-objective optimization problem and derive selective conflict-aware position updates that reduce interference between competing objectives while preserving the contribution of each loss. We further observe that gradient conflicts persist throughout training and are most severe in Gaussian position parameters, indicating that objective conflict is closely tied to geometry optimization. We therefore treat position-gradient conflict as a geometry-aware signal to guide densification and pruning. Experiments show that our method achieves better rendering performance while improving representation compactness and rendering efficiency over existing 3DGS variants.
\end{abstract}

%% file: sec/1_intro.tex
\section{Introduction}
\label{sec:intro}

\begin{figure}[!t]
    \centering
    \includegraphics[width=\columnwidth]{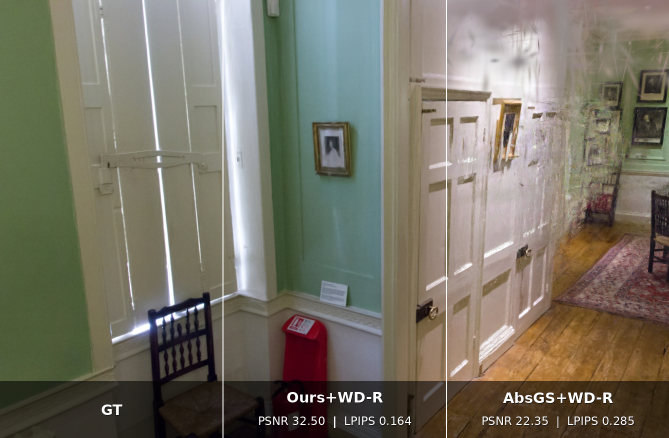}
    \caption{{\textbf{Comparison after introducing the perceptual WD-R loss \cite{dropin}.}
Other methods produce severe artifacts, whereas our method remains stable and improves both reconstruction fidelity (PSNR) and perceptual quality (LPIPS).} }
    \label{fig:intro}
\end{figure}
Neural rendering and novel view synthesis (NVS) have been long-standing research topics in computer vision and graphics, with growing applications in robotics, virtual and augmented reality, and autonomous-driving\cite{maniGaussian,srbfGaussian,icp3DGS}. Neural Radiance Fields (NeRFs)~\cite{mildenhall2020nerf} represent scenes as implicit radiance fields and synthesize novel views through volumetric rendering, but their expensive ray tracing makes real-time rendering difficult. More recently, 3D Gaussian Splatting (3DGS)~\cite{kerbl20233d} has emerged as a powerful alternative. It represents scenes using explicit anisotropic Gaussian primitives and employs differentiable rasterization to achieve high-quality rendering at near-real-time speed. Beyond its efficient rasterizer, this performance relies critically on jointly optimizing Gaussian attributes and adaptively allocating primitives throughout the scene.

The original 3DGS optimizes a weighted combination of $L_1$ and D-SSIM losses and uses accumulated view-space positional gradients to control primitive densification. Although simple and effective, the recorded gradient magnitude does not always faithfully reflect local representation demands. Subsequent studies\cite{absGS,frequencyADC,gaussianMCMC,pixelGS,perceptualGS} have tried to improve the optimization and densification of
3DGS from various perspectives. However, they largely overlook a
fundamental aspect of its training objective, that vanilla 3DGS jointly optimizes two distinct losses: while the $L_1$ loss promotes pixel-level fidelity between the rendered image and ground truth, D-SSIM encourages local structural consistency. As noted in prior work~\cite{zhao2017loss},
these objectives favor different image characteristics and exhibit
complementary optimization behaviors, potentially inducing competing
updates to shared Gaussian parameters. Their joint optimization, therefore, naturally resembles a multi-objective optimization (MOO) problem. Moreover, as
recent 3DGS applications increasingly prioritize human-perceived quality \cite{vrSplatting,vrSplat,perceptualGS},
perceptual losses, such as LPIPS loss and Wasserstein distance (WD-R), have been incorporated as additional training
objectives~\cite{dropin}. The greater heterogeneity among these objectives
makes cross-objective gradient conflict increasingly consequential.

Unlike conventional MOO, where gradient conflict is typically analyzed over globally shared parameters, the explicit and spatially localized representation of 3DGS makes it possible to address this problem at the primitive level. We observe that gradient-direction conflicts are mainly concentrated in Gaussian position attributes, and high-conflict Gaussians are strongly correlated with regions of large rendering error. This suggests that position-gradient conflict carries meaningful geometric information: it may indicate either insufficient local representation capacity or inaccurate Gaussian geometry.

Based on these observations, we propose \textbf{GC-GS}, a
gradient-conflict-aware 3D Gaussian Splatting framework. GC-GS
measures directional disagreement between objective-specific gradients for individual primitives, selectively adjusts conflicting Gaussian position updates, and uses persistent conflict to guide density control by densifying regions that require additional representation capacity while suppressing primitives with unreliable geometry. GC-GS introduces no additional loss terms and is compatible with diverse training objectives, including the conventional $L_1+$D-SSIM loss
and perceptual objectives such as LPIPS-based composite loss and WD-R \cite{dropin}. Our contributions are summarized as follows:
\begin{itemize}
    \item We identify cross-objective gradient conflict as a spatially heterogeneous, primitive-level phenomenon in 3DGS.
    \item We introduce selective geometry-level gradient reconciliation and conflict-aware
density control to coordinate heterogeneous objectives and guide the
distribution of Gaussian primitives.
    \item Experiments across multiple datasets and perceptual objectives demonstrate that our framework achieves consistent performance improvements with better rendering efficiency.
\end{itemize}

%% file: sec/2_related_work.tex
\section{Related Work}
\label{sec:Related}

\subsection{3D Gaussian Splatting}
Extensive efforts have been made to address the limitations of vanilla 3D Gaussian
Splatting~\cite{kerbl20233d} from complementary perspectives.

\noindent\textbf{3DGS optimization.}
Several methods \cite{mipSplatting,analyticSplatting,stopThePop} focus on improving reconstruction quality or accelerating 3DGS
optimization without fundamentally changing the framework.
For example, Mip-Splatting~\cite{mipSplatting} introduces spatial filtering to reduce
aliasing, while StopThePop~\cite{stopThePop} improves the Gaussian sorting for better view consistency. Other works explore more advanced backpropagation strategies or more
structured Gaussian representations~\cite{threeDGSLM,threeDGS2,taming3DGS,twoDGS,gof,scaffoldGS}.

\noindent\textbf{3DGS densification.}
Another line of research improves how Gaussian primitives are allocated. 
Pixel-GS~\cite{pixelGS} and Revising-Densification~\cite{revisingDensification} use pixel coverage and
reconstruction error to refine densification decisions, while
FADC~\cite{frequencyADC} and
Perceptual-GS~\cite{perceptualGS} introduce frequency and human perceptual
cues as additional information to guide the densification process. From the gradient perspective, AbsGS~\cite{absGS} mitigates
pixel-level sub-gradient cancellation, GDAGS~\cite{gdags} models
directional consistency between pixels, and SteepGS~\cite{steepGS} derives when and
along which direction a Gaussian should be split. Nevertheless, these
methods either introduce additional density cues or analyze gradients
after the training objectives have been aggregated, leaving
cross-objective interactions unexplored.

\noindent\textbf{3DGS pruning.}
Apart from controlling densification, a complementary line of work
removes redundant primitives to improve representation efficiency. Some
methods aim to construct compact Gaussian representations with fewer
primitives~\cite{miniSplatting,pup3DGS,speedySplat,trimmingFat}.
Specifically, Mini-Splatting~\cite{miniSplatting} combines simplification
with importance-based sampling, while PUP 3D-GS~\cite{pup3DGS} estimates
Gaussian importance through second-order reconstruction sensitivity.
Other methods target the final model size by combining pruning with
attribute compression~\cite{compact3DGS,lightGaussian,mesonGS,sizeGS}.
Despite their different efficiency objectives, existing
pruning criteria easily lead to degraded rendering
quality under aggressive pruning.

\subsection{Multi-Objective Optimization}

Multi-objective optimization seeks a shared solution for multiple
objectives that generally cannot be minimized simultaneously. Existing
gradient-based methods, such as MGDA~\cite{mgda},
PCGrad~\cite{pcgrad}, and CAGrad~\cite{cagrad}, construct or modify a
shared update direction to mitigate competition among objectives.
3DGS naturally involves such interactions: its conventional objective
combines $L_1$ and D-SSIM, while recent methods further introduce
perceptual losses such as LPIPS and WD-R~\cite{dropin}. These objectives
emphasize different aspects of reconstruction, including pixel fidelity,
local structure, and perceptual texture, and may therefore produce
conflicting gradients. Although gradient conflict has been 
studied in multi-task learning, interactions among different reconstruction
objectives remain largely unexplored in 3DGS.

%% file: sec/3_method.tex
\section{Method}
\label{sec:Method}


\subsection{Preliminary}

3D Gaussian Splatting (3DGS)~\cite{kerbl20233d} represents a scene using
anisotropic Gaussian primitives
$\mathcal{G}=\{\mathcal{G}_i(\boldsymbol{\mu}_i,
\boldsymbol{\Sigma}_i,\alpha_i,\mathbf{c}_i)\}_{i=1}^{N}$, where
$\boldsymbol{\mu}_i$, $\boldsymbol{\Sigma}_i$, $\alpha_i$, and
$\mathbf{c}_i$ denote position, covariance, opacity, and spherical
harmonic coefficients, respectively. The covariance is parameterized as
\begin{equation}
    \boldsymbol{\Sigma}_i
    =
    \mathbf{R}_i\mathbf{S}_i\mathbf{S}_i^\top\mathbf{R}_i^\top,
\end{equation}
where $\mathbf{R}_i$ and $\mathbf{S}_i$ represent rotation and scale.
For a viewpoint $v$, the Gaussians are projected onto the image plane,
sorted by depth, and alpha-blended as
\begin{equation}
    \hat{\mathbf C}_v(\mathbf p)
    =
    \sum_i
    \mathbf c_i(v)\,
    \tilde{\alpha}_{i,v}(\mathbf p)
    \prod_{j<i}\left(1-\tilde{\alpha}_{j,v}(\mathbf p)\right),
\end{equation}
where
$\tilde{\alpha}_{i,v}(\mathbf p)=
\alpha_i\mathcal{G}'_{i,v}(\mathbf p)$ is the effective opacity of the
projected Gaussian on the image plane.
The original 3DGS is optimized using the photometric loss:
\begin{equation}
    \mathcal L
    =
    (1-\lambda)\mathcal L_1
    +
    \lambda\mathcal L_{\mathrm{D\text{-}SSIM}}.
\end{equation}
Adaptive density control accumulates the view-space positional-gradient
magnitude of each Gaussian:
\begin{equation}
    a_i
    =
    \frac{1}{|\mathcal V_i|}
    \sum_{v\in\mathcal V_i}
    \left\|
    \frac{\partial\mathcal L_v}
    {\partial\boldsymbol{\mu}^{2D}_{i,v}}
    \right\|_2.
\end{equation}
Gaussians with $a_i$ above a threshold are cloned when small and split
when large, while low-opacity primitives are periodically removed.

More generally, for
$\mathcal L=\sum_{k=1}^{K}\lambda_k\mathcal L_k$, the gradient of
Gaussian $i$ is
\begin{equation}
    \mathbf g_i
    =
    \sum_{k=1}^{K}\lambda_k\mathbf g_{k,i},
    \qquad
    \mathbf g_{k,i}
    =
    \nabla_{\boldsymbol{\theta}_i}\mathcal L_k.
\end{equation}
Conventional 3DGS uses only the aggregated gradient $\mathbf g_i$,
without explicitly considering directional conflict among its
objective-specific components.

\subsection{Overview}
\begin{figure}[t]
    \centering
    \begin{minipage}[t]{0.49\linewidth}
        \centering
        \includegraphics[width=\linewidth]{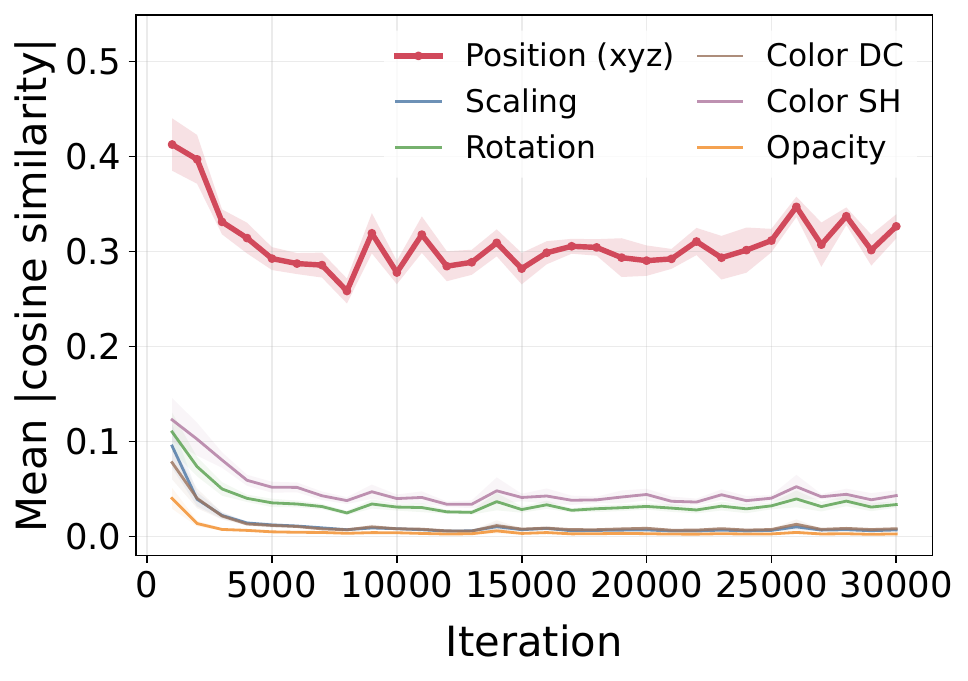}
        \vspace{1mm}
        \parbox{\linewidth}{\centering\small (a) Mean conflict of different attributes during training}
    \end{minipage}
    \hfill
    \begin{minipage}[t]{0.49\linewidth}
        \centering
        \includegraphics[width=\linewidth]{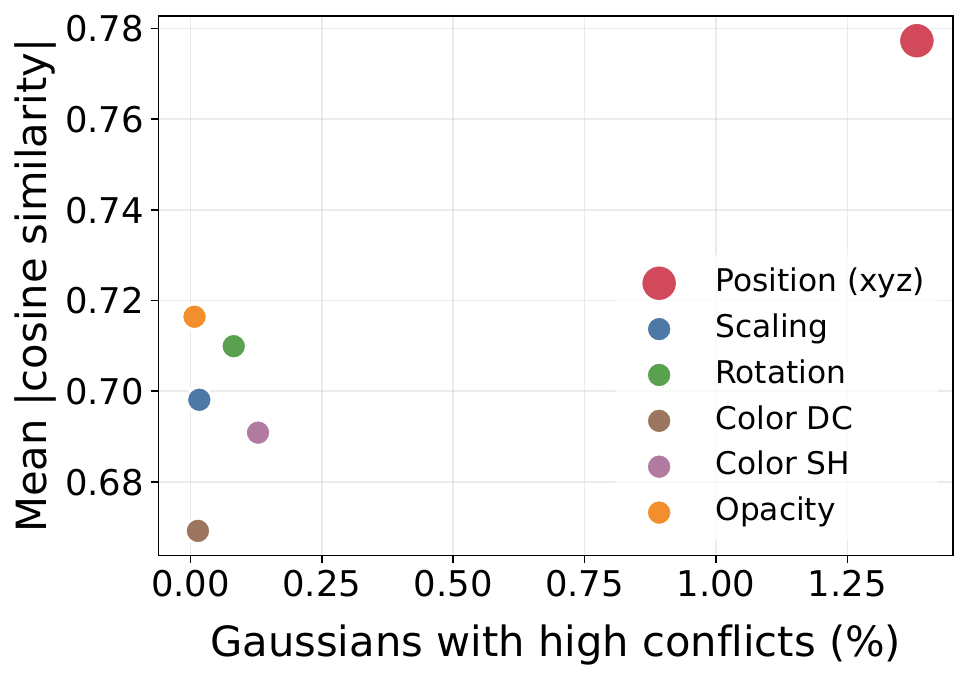}
        \vspace{1mm}
        \parbox{\linewidth}{\centering\small (b) Mean conflict for high-conflict Gaussians}
    \end{minipage}

    \vspace{1.5mm}

    \begin{minipage}[t]{0.49\linewidth}
        \centering
        \includegraphics[width=\linewidth]{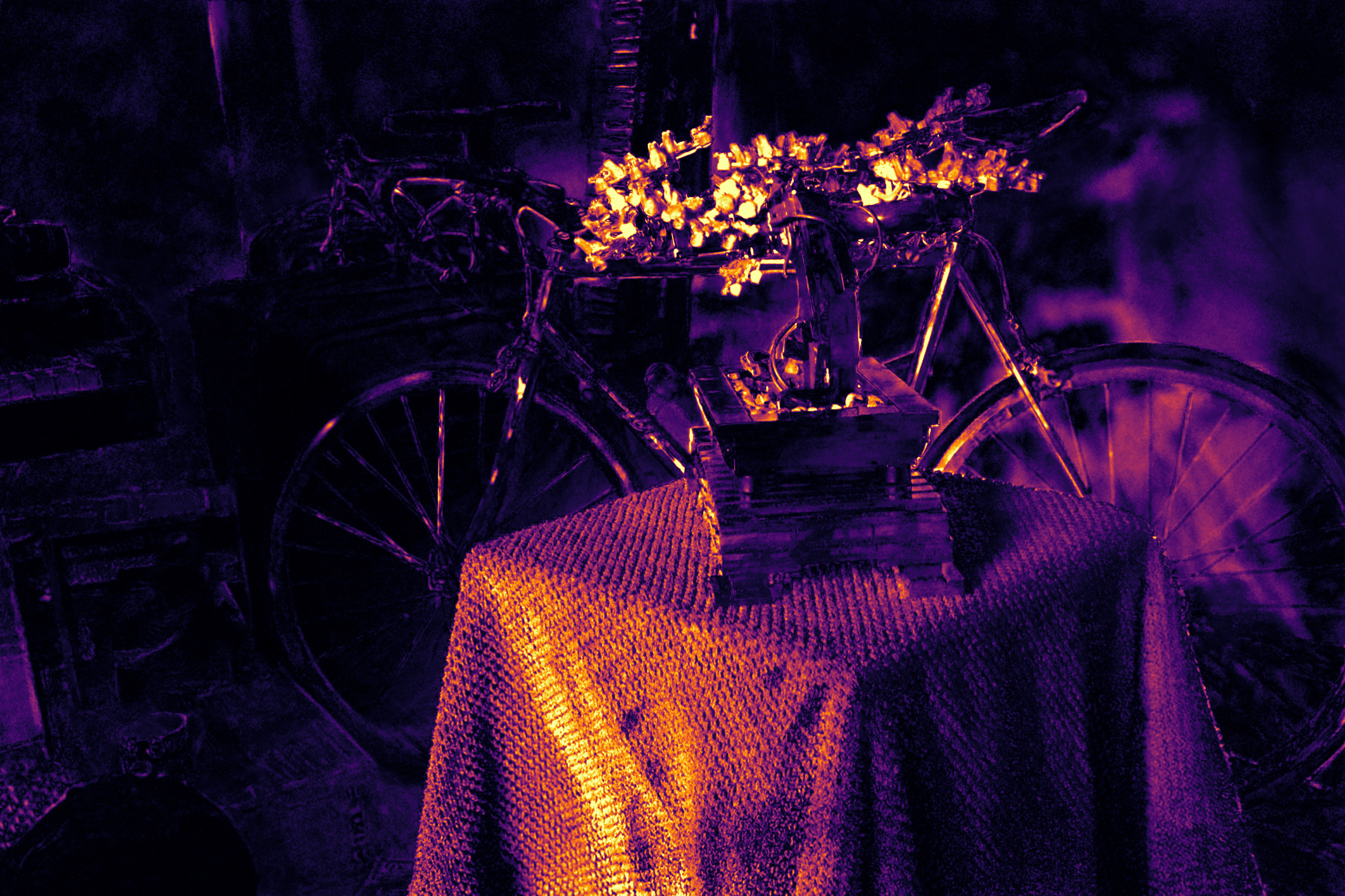}
        \vspace{1mm}
        \parbox{\linewidth}{\centering\small (c) L1 error map}
    \end{minipage}
    \hfill
    \begin{minipage}[t]{0.49\linewidth}
        \centering
        \includegraphics[width=\linewidth]{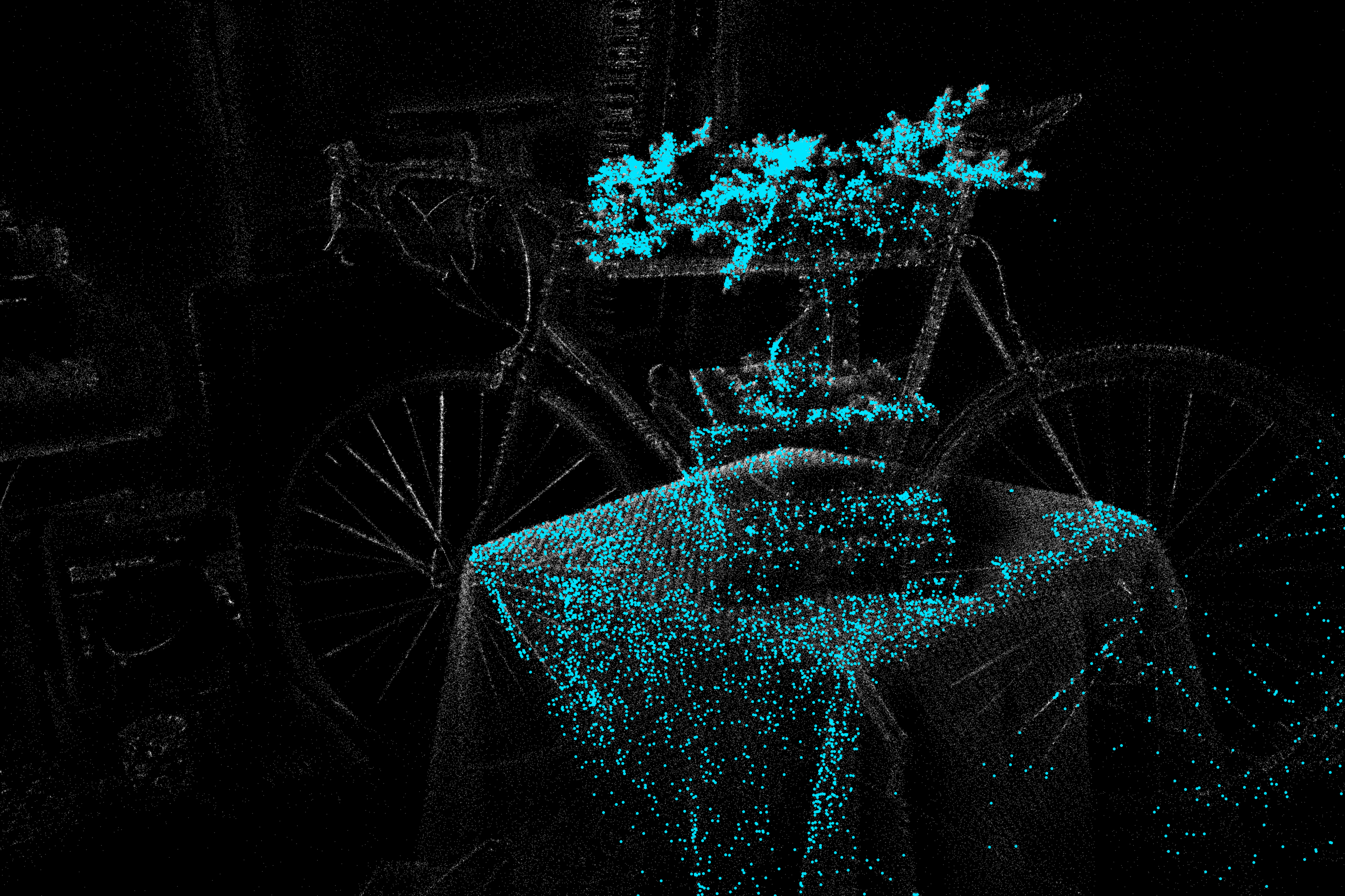}
        \vspace{1mm}
        \parbox{\linewidth}{\centering\small (d) High-conflict Gaussian positions}
    \end{minipage}

    \caption{
    Attribute-wise gradient conflict in vanilla 3DGS.
    Results in (a) and (b) are averaged over all nine scenes of the
    Mip-NeRF 360 dataset~\cite{barron2022mipnerf360}.
    Visualizations in (c) and (d) are from the \textit{Bonsai} scene in
    Mip-NeRF 360, showing that high-conflict Gaussians tend to appear in
    challenging regions with large reconstruction error.
    }
    \label{fig:attribute_conflict_analysis}
\end{figure}
As shown in Fig.~\ref{fig:attribute_conflict_analysis}(a) and (b), we
observe that during vanilla 3DGS training, position-gradient conflicts
persist throughout optimization, whereas conflicts in other Gaussian
attributes largely diminish as training converges. Gaussian positions are
also substantially more likely to exhibit severe objective conflict.
Moreover, high-conflict Gaussian positions spatially overlap with
high-error rendering regions, as illustrated in
Fig.~\ref{fig:attribute_conflict_analysis}(c) and (d). We therefore apply selective position gradient reconciliation to
high-conflict primitives, adjusting their updates only when the
objective-specific position gradients strongly disagree.
Beyond adjusting the update direction, we further treat persistent conflict
as a geometry-related signal. During early training, high conflict often
indicates insufficient local capacity, where the current Gaussian
allocation cannot simultaneously satisfy both objectives. Near
convergence, persistent conflict instead suggests unstable or redundant
geometry that continues to receive inconsistent supervision. Based on
this stage-dependent interpretation, we introduce conflict-guided
densification and pruning to adapt the Gaussian distribution throughout
training.

\subsection{Geometry-Level Gradient Reconciliation}
\label{subsec:geometry_level_gradient_harmonization}
\begin{figure*}[t]
    \centering
    \includegraphics[width=\textwidth]{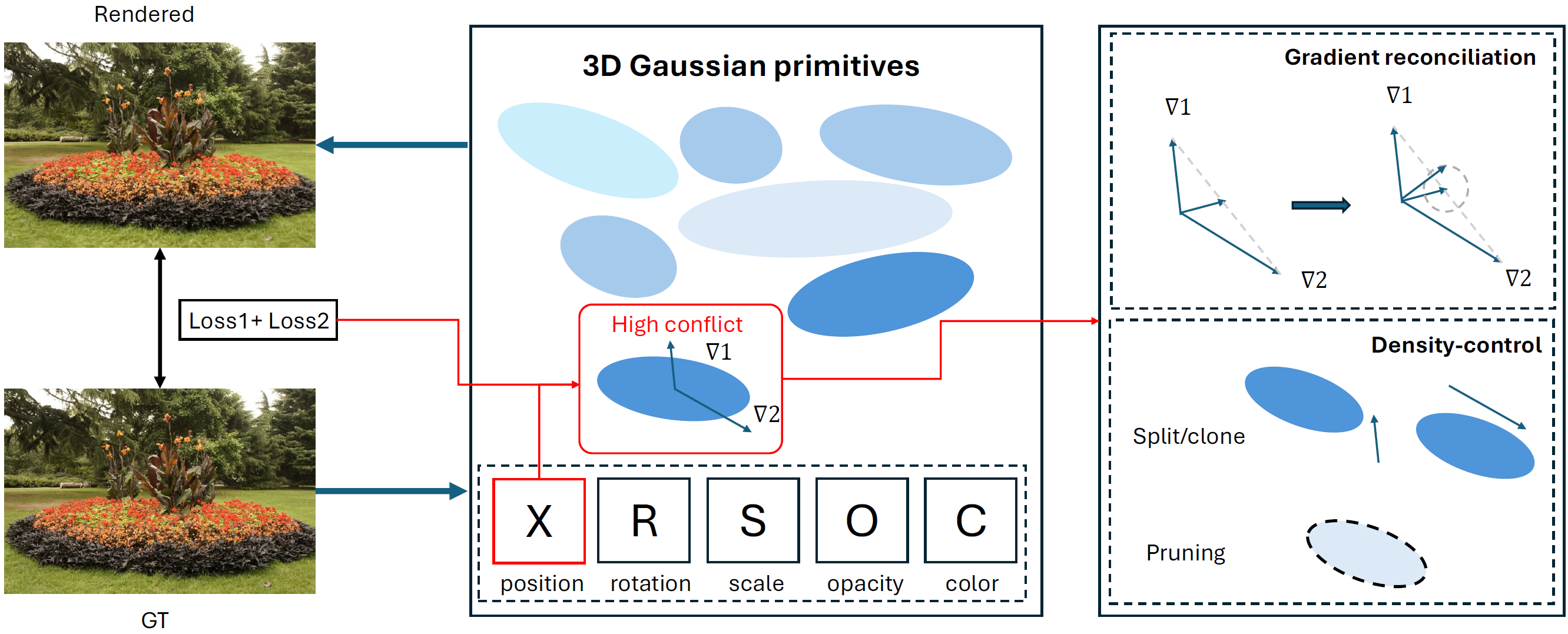}
    \caption{
    Overview of GC-GS. Given two training objectives, GC-GS first measures
primitive-level conflicts from their position gradients. High-conflict
Gaussians receive a selective conflict-aware position gradient adjustment, while
well-aligned Gaussians follow the original update. The persistent conflict
signal is further used for density control: in the early stage, it
encourages splitting or cloning to increase local
representation capacity; in the late stage, it suppresses high-conflict
and high-error Gaussians.
    }
    \label{fig:main_figure}
\end{figure*}

Let $\mathcal{L}_a$ and $\mathcal{L}_b$ denote the two training
objectives. Depending on the training configuration, we use
\begin{equation}
(\mathcal{L}_a,\mathcal{L}_b)=
\begin{cases}
(\mathcal{L}_{1},\mathcal{L}_{\mathrm{D\text{-}SSIM}}),
& \text{original},\\
(\mathcal{L}_{\mathrm{original}},\mathcal{L}_{\mathrm{WD}}),
& \text{WD-R},\\
(\mathcal{L}_{\mathrm{original}},\mathcal{L}_{\mathrm{LPIPS}}),
& \text{LPIPS composite}.
\end{cases}
\end{equation}
After absorbing their loss coefficients, their joint optimization is
formulated as
\begin{equation}
    \min_{\boldsymbol{\Theta}}
    \left\{
    \mathcal{L}_a(\boldsymbol{\Theta}),
    \mathcal{L}_b(\boldsymbol{\Theta})
    \right\},
\end{equation}
which naturally forms a multi-objective optimization problem.

Unlike neural networks, whose shared parameters globally affect many
predictions, 3DGS consists of spatially localized primitives. Different
Gaussians can therefore exhibit substantially different objective
preferences. Applying a single global-level adjustment imposes the same
trade-off on the entire scene and may unnecessarily alter Gaussians whose
objectives are already compatible. Inspired by gradient-based
MOO~\cite{mgda,cagrad}, we instead resolve objective conflict selectively
at the individual Gaussian level.

We focus on Gaussian positions because their gradients directly determine
both geometry optimization and adaptive density control. For each visible
Gaussian $G_i$ with center $\boldsymbol{\mu}_i\in\mathbb{R}^3$, we compute
\begin{equation}
    \boldsymbol{g}_i^k
    =
    \nabla_{\boldsymbol{\mu}_i}\mathcal{L}_k,
    \qquad k\in\{a,b\}.
\end{equation}
A conflict occurs when the angle between the two gradients exceeds
$90^\circ$, such that
\begin{equation}
    \left\langle
    \boldsymbol{g}_i^a,\boldsymbol{g}_i^b
    \right\rangle < 0.
\end{equation}
In this case, descending along one gradient may oppose the improvement
preferred by the other objective. 

For a conflicting Gaussian, let
\begin{equation}
    \bar{\boldsymbol{g}}_i
    =
    \frac{1}{2}
    \left(
    \boldsymbol{g}_i^a+\boldsymbol{g}_i^b
    \right)
\end{equation}
denote its original averaged gradient. We seek a position update that
maximizes the improvement of the least-favored objective while remaining
close to the original optimization direction:
\begin{equation}
\begin{aligned}
    \boldsymbol{d}_i^*
    ={}&
    \arg\max_{\boldsymbol{d}_i}
    \min_{k\in\{a,b\}}
    \left\langle
    \boldsymbol{g}_i^k,\boldsymbol{d}_i
    \right\rangle,\\
    &\mathrm{s.t.}\quad
    \|\boldsymbol{d}_i-\bar{\boldsymbol{g}}_i\|_2
    \leq c\|\bar{\boldsymbol{g}}_i\|_2,
\end{aligned}
\end{equation}
where $c$ controls the reconciliation strength. The resulting position
gradient is
\begin{equation}
    \widehat{\boldsymbol{g}}_i=
    \begin{cases}
        \boldsymbol{d}_i^*,
        & \langle\boldsymbol{g}_i^a,\boldsymbol{g}_i^b\rangle<0,\\
        \bar{\boldsymbol{g}}_i,
        & \text{otherwise}.
    \end{cases}
\end{equation}
The constrained problem reduces to a convex one-dimensional optimization
and is efficiently solved by bisection. 
Only the position gradients of conflicting Gaussians are replaced.
Compatible Gaussian positions and all remaining attributes, including
scale, rotation, opacity, and color, retain their original aggregated
gradients. This selective design prevents either objective from dominating
high-conflict primitives without disturbing the optimization of the
remaining scene.
\subsection{Conflict-Guided Density Control}

\subsubsection{Windowed Objective Conflict Measurement}

\input{Figure/qualitative_results_original}

To measure the conflict severity of each Gaussian directly in 3D space,
we accumulate its position-gradient disagreement over the ten most
recently sampled training views before each density-control operation.
For Gaussian $G_i$ observed at view $t$, the position gradient induced by
objective $k$ is defined as
\begin{equation}
    \boldsymbol{p}_{i,t}^{k}
    =
    \frac{\partial \mathcal{L}_{k}^{t}}
    {\partial \boldsymbol{\mu}_{i}},
    \qquad k\in\{a,b\},
\end{equation}
where $\boldsymbol{\mu}_i\in\mathbb{R}^3$ is the center of $G_i$, and
$a$ and $b$ denote the two optimization objectives.

We measure conflict using only the angle between the two objective-specific
position gradients:
\begin{equation}
    c_{i,t}
    =
    \max\left(
    -
    \frac{
    \left\langle
    \boldsymbol{p}_{i,t}^{a},
    \boldsymbol{p}_{i,t}^{b}
    \right\rangle}
    {
    \|\boldsymbol{p}_{i,t}^{a}\|_2
    \|\boldsymbol{p}_{i,t}^{b}\|_2+\epsilon
    },
    0
    \right).
\end{equation}
This score lies in $[0,1]$, where $0$ indicates non-conflicting gradients
and $1$ indicates completely opposite directions.

To avoid treating weak noisy gradients as meaningful conflicts, we only
activate the score when both objective-specific gradient magnitudes are
larger than the $75$th percentile. The windowed conflict score for $G_i$
is then defined as
\begin{equation}
    C_i^{\mathrm{win}}
    =
    \frac{1}{|\mathcal W|}
    \sum_{t\in\mathcal W}v_{i,t}c_{i,t},
\end{equation}
Here, $\mathcal W$ is the recent-view window, $c_{i,t}$ is the angular
conflict score, and $v_{i,t}$ is the activation mask after the 
gradient-magnitude selection. The resulting
score captures persistent angular disagreement between the two objectives
while filtering out conflicts caused by weak gradients.

\subsubsection{Conflict-Guided Directional Densification}

During the early stage of training, we interpret high conflict as an
indication that the local representation capacity of a Gaussian is
insufficient to accommodate the optimization preferences of both
objectives. We therefore encourage such Gaussians to densify by adding
them to the candidates selected by the original densification process:
\begin{equation}
    h_i^{\mathrm{den}}
    =
    \mathbb{I}[G_i^{\mathrm{ADC}}>\tau_g]
    \lor
    \mathbb{I}[C_i^{\mathrm{win}}>\tau_c^{\mathrm{den}}],
\end{equation}
where $G_i^{\mathrm{ADC}}$ and $\tau_g$ denote the original densification
gradient statistic and threshold, while $\tau_c^{\mathrm{den}}$ is the
conflict threshold.

At each densification step, the original scale criterion determines
whether a selected Gaussian is cloned or split. For candidates introduced
by the conflict criterion, we perform conflict-aware initialization by
placing the resulting pair along the current descent directions of the two
objectives. This operation expands the local representation along both
objective-specific descent directions instead of forcing a single
primitive to follow their conflicting average. This two-direction placement
can be interpreted as encouraging descent for both objectives.

\subsubsection{Conflict-Guided Opacity Suppression}

\input{tables/results}

\input{tables/efficiency}

During the later stage of training, the scene has already been allocated
a sufficient number of Gaussians through early densification. We therefore
interpret Gaussians that still exhibit persistent conflict as more likely
to contain noisy or incorrectly optimized geometry, which prevents the two
objectives from providing compatible supervision. Instead of further
increasing their capacity, we reduce their rendering contribution to
limit the propagation of erroneous information.

To avoid suppressing Gaussians based on conflict alone, we additionally
measure their accumulated contribution to high-error regions across the
same multi-view window:
\begin{equation}
    R_i^{\mathrm{win}}
    =
    \frac{
    \sum_{t\in\mathcal W}
    \sum_p \omega_i^t(p)E^t(p)}
    {|\mathcal W|},
\end{equation}
where $\omega_i^t(p)$ denotes the alpha-blending contribution of $G_i$ to
pixel $p$, and $E^t(p)$ is the corresponding rendering-error map. A
Gaussian is selected only when both its conflict and error contribution
exceed their respective thresholds.

We then directly attenuate its opacity:
\begin{equation}
    \alpha_i
    \leftarrow
    \alpha_i\exp(-\lambda_p C_i^{\mathrm{win}}),
\end{equation}
where $\lambda_p$ is a fixed modulation coefficient and is not optimized.
If the selected Gaussian provides useful rendering information, subsequent
reconstruction gradients can gradually recover its opacity. Otherwise,
repeated suppression keeps its opacity low, allowing it to be removed by
the original opacity-based pruning mechanism of 3DGS.

%% file: Figure/qualitative_results_original.tex
\begin{figure*}[!ht]
    \centering
    \includegraphics[width=\textwidth]{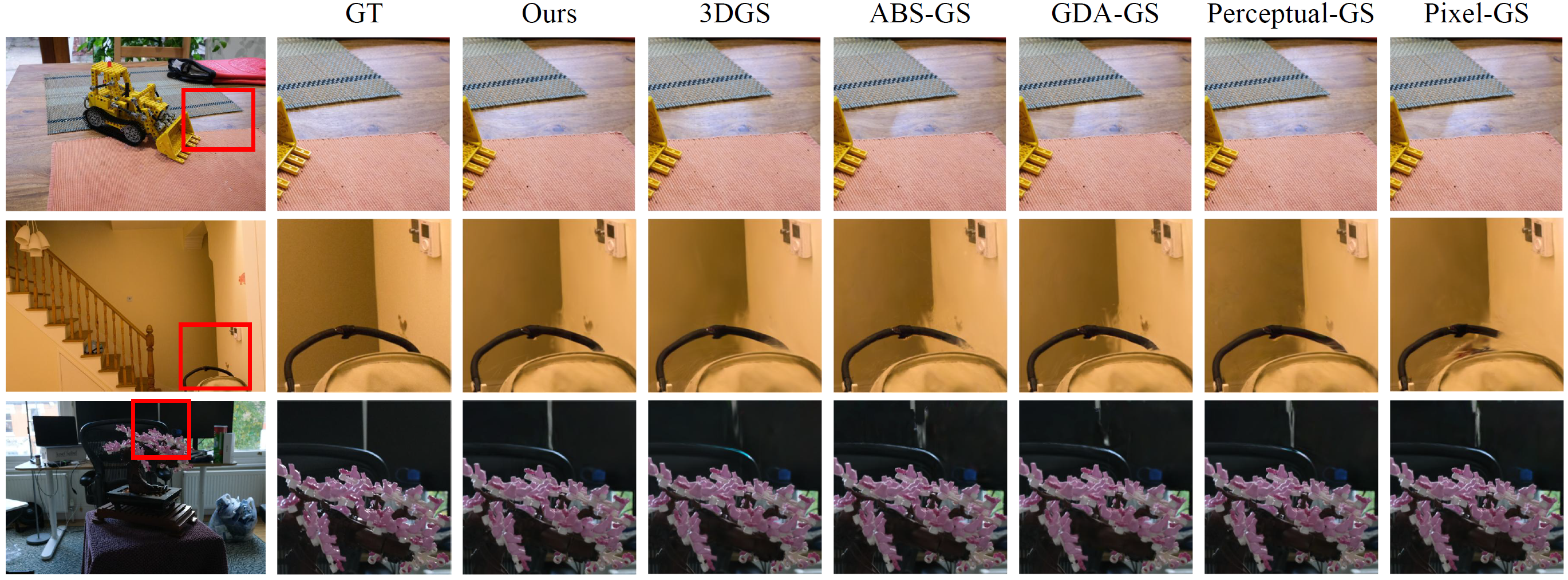}
    \caption{
Qualitative comparison under the original 3DGS loss. GC-GS better
captures lighting effects, preserves details, and suppresses artifacts.
}
    \label{fig:qualitative_results_original}
\end{figure*}

%% file: tables/results.tex
\begin{table*}[ht]
\centering
\small
\caption{
Quantitative reconstruction results on three datasets with original training loss.
We report PSNR$\uparrow$, SSIM$\uparrow$, LPIPS$\downarrow$, and \#G$\downarrow$, where \#G denotes the Gaussian count in millions (M).
The \colorbox{best}{best} and
\colorbox{second}{second-best} results are highlighted.
}
\label{tab:quantitative_results}

\setlength{\tabcolsep}{3pt}
\renewcommand{\arraystretch}{1.08}

\begin{tabular*}{\textwidth}{@{\extracolsep{\fill}}lcccccccccccc@{}}
\toprule
Method
& \multicolumn{4}{c}{Mip-NeRF 360}
& \multicolumn{4}{c}{Tanks \& Temples}
& \multicolumn{4}{c}{Deep Blending} \\
\cmidrule(lr){2-5}
\cmidrule(lr){6-9}
\cmidrule(lr){10-13}

& PSNR$\uparrow$
& SSIM$\uparrow$
& LPIPS$\downarrow$
& \#G (M)$\downarrow$
& PSNR$\uparrow$
& SSIM$\uparrow$
& LPIPS$\downarrow$
& \#G (M)$\downarrow$
& PSNR$\uparrow$
& SSIM$\uparrow$
& LPIPS$\downarrow$ 
& \#G (M)$\downarrow$\\
\midrule

3DGS

& 27.82 & 0.827 & 0.201 & \cellcolor{second}2.54
& 23.73 & 0.852 & 0.171 & 1.55
& 29.85 & 0.907 & 0.238 & 2.51 \\

Pixel-GS

& 27.86
& 0.835
& \cellcolor{second}0.176
& 5.25
& 23.76
& 0.856
& \cellcolor{best}0.149
& 4.52
& 28.94
& 0.897
& 0.248 
& 4.64 \\

Perceptual-GS

& \cellcolor{second}27.96
& \cellcolor{second}0.839
& \cellcolor{best}0.173
& 2.69
& \cellcolor{second}23.77
& \cellcolor{second}0.857
& \cellcolor{second}0.150
& 1.73
& 29.90
& 0.906
& \cellcolor{best}0.231 & 2.90 \\

AbsGS

& 27.46 & 0.812 & 0.191
& 3.37
& 23.76 & 0.854 & 0.162
& \cellcolor{best}1.30
& 29.60 & 0.901 & 0.237 & \cellcolor{second}1.94 \\

GDAGS

& 27.71 & 0.833 & 0.179
& 2.81
& 23.50 & 0.849 & 0.172
& 1.34
& \cellcolor{second}29.98
& \cellcolor{second}0.908
& \cellcolor{second}0.231 & 2.04 \\

Ours

& \cellcolor{best}28.12
& \cellcolor{best}{0.839}
& 0.180
& \cellcolor{best}{2.27}
& \cellcolor{best}{24.01}
& \cellcolor{best}{0.860}
& 0.161
& \cellcolor{second}1.33
& \cellcolor{best}{30.07}
& \cellcolor{best}{0.911}
& 0.239 & \cellcolor{best}{1.71}\\

\bottomrule
\end{tabular*}

\end{table*}

%% file: tables/efficiency.tex
\begin{table}[!ht]
\centering
\small
\caption{
Efficiency comparison across all three datasets. Rendering speed is reported in FPS, and training time is reported in minutes. Our method achieves the highest average rendering speed while incurring moderate training overhead.
}
\label{tab:efficiency_results}

\setlength{\tabcolsep}{2.5pt}
\renewcommand{\arraystretch}{1.05}

\resizebox{\columnwidth}{!}{
\begin{tabular}{lcccccc}
\toprule
Method
& \multicolumn{2}{c}{Mip-NeRF 360}
& \multicolumn{2}{c}{Tanks\&Temples}
& \multicolumn{2}{c}{Deep Blending} \\
\cmidrule(lr){2-3}
\cmidrule(lr){4-5}
\cmidrule(lr){6-7}
& FPS$\uparrow$ & Time$\downarrow$
& FPS$\uparrow$ & Time$\downarrow$
& FPS$\uparrow$ & Time$\downarrow$ \\
\midrule
3DGS          & 209.00 & 16.68 & 238.97 & 9.77  & 187.91 & 17.42 \\
Pixel-GS      & 125.61 & 20.06 & 106.99 & 15.07 & 131.42 & 20.32 \\
Perceptual-GS & 193.90 & 23.70 & 223.41 & 13.46 & 197.77 & 23.38 \\
AbsGS         & 222.79 & 15.51 & 314.27 & 7.26  & 304.97 & 11.92 \\
GDAGS         & 240.05 & 13.24 & 292.17 & 6.91  & 315.06 & 11.03 \\
Ours          & 267.07 & 19.98  & 284.39 &  14.28   & 341.56 & 16.65    \\
\bottomrule
\end{tabular}}
\end{table}

%% file: sec/4_experiment.tex
\section{Experiment}
\label{sec:Experiment}
\subsection{Experiment Setting}

\noindent\textbf{Evaluation datasets.}
Following the standard 3DGS evaluation protocol, we conduct experiments
on 13 scenes from three widely used datasets. Specifically, we use all
nine scenes from Mip-NeRF 360~\cite{barron2022mipnerf360}, the
\textit{Train} and \textit{Truck} scenes from Tanks \& 
Temples~\cite{knapitsch2017tanks}, and the \textit{Playroom} and
\textit{DrJohnson} scenes from Deep Blending~\cite{hedman2018deepblending}.
Together, these datasets cover indoor and outdoor environments with
diverse scene scales, geometric structures, and appearance complexity.
We follow the official training and evaluation splits for all scenes.

\noindent\textbf{Baselines.}
We compare our method, GC-GS, with the original 3DGS~\cite{kerbl20233d} and recent
representative 3DGS-based methods that improve gradient aggregation or
adaptive density control. These include Pixel-GS~\cite{pixelGS},
AbsGS~\cite{absGS}, Perceptual-GS~\cite{perceptualGS}, and
GDAGS~\cite{gdags}. These baselines cover pixel-coverage-aware,
absolute-gradient-based, perceptually guided, and
gradient-direction-aware densification strategies, enabling a
comprehensive evaluation of both reconstruction quality and Gaussian
allocation.

\noindent\textbf{Metrics.}
We report PSNR, SSIM~\cite{wang2004ssim}, and
LPIPS~\cite{zhang2018lpips} as standard novel-view synthesis metrics.
For experiments focusing on perceptual quality, we additionally report
DISTS~\cite{ding2020dists}. To assess efficiency, we report the
number of Gaussian primitives (\#G), rendering speed in frames per second
(FPS), and the total training time in minutes. All FPS and training-time measurements
are obtained using the same hardware and evaluation settings.

\noindent\textbf{Implementation Details.}
Our method is implemented on the official open-source codebase of vanilla
3DGS, with all shared hyperparameters kept unchanged. Training runs for
30K iterations. Densification is disabled after 15K iteration,
and conflict-aware pruning is performed every 3K iterations until
training terminates at iteration 30K. The conflict score is computed over
a window of the ten most recently sampled training views. The gradient control parameter $c$ is set to 0.2, the error threshold to 0.1 and the modulation coefficient $\lambda_p$ to 0.4. All experiments are conducted on a single NVIDIA H100 GPU. For fair
comparison, all competing methods are reproduced using their official
open-source implementations with their default hyperparameter settings.

\input{tables/perceptual_loss_results}

\input{Figure/quanlitative_results_wdr}
\input{Figure/quanlitative_results_lpips}

\subsection{Comparison with  State-of-the-Art Methods}
\noindent\textbf{Comparison on original loss.}
We report the detailed quantitative compa  rison of GC-GS with state-of-the-art methods across PSNR, SSIM, LPIPS, and the Gaussian count in Table~\ref{tab:quantitative_results}. Notably, our method achieves the best PSNR and SSIM scores across all datasets while maintaining the fewest or second-fewest Gaussians, indicating improvements in both reconstruction quality and representation efficiency. This supports that the gradient conflict between loss components can provide useful supervisory signals during training. By explicitly exploiting these conflicts, GC-GS can better coordinate the contributions of individual objectives and guide the optimization toward a more effective solution. As depicted in Figure~\ref{fig:qualitative_results_original}, GC-GS reconstructs scenes with high detail fidelity, including enhanced surface reflectance, sharpened edges, and clear boundaries, mitigating artifacts observed in other methods.

\noindent\textbf{Comparison on perceptual loss.}
We further conducted experiments by incorporating perceptual objectives, i.e., WD-R and LPIPS~\cite{dropin}, into the original D-SSIM and L1 losses, as described in Section~\ref{subsec:geometry_level_gradient_harmonization}. The quantitative results are reported in Table~\ref{tab:quantitative_perceptual_loss_results}, while the qualitative results for WD-R and LPIPS are shown in Figures~\ref{fig:qualitative_results_wdr} and~\ref{fig:qualitative_results_lpips}, respectively.
Across all evaluated datasets and under both perceptual-loss settings, our method consistently matches or outperforms the baselines in both fidelity metrics, including PSNR and SSIM, and perceptual-quality metrics, including LPIPS and DISTS. The improvement is particularly notable on Mip-NeRF 360. When WD-R is added, our method improves PSNR by 0.37 over vanilla 3DGS while reducing LPIPS by 0.014, demonstrating that gradient conflict guidance can effectively facilitate the joint optimization of fidelity-oriented and perception-oriented objectives.
It is worth noting that perceptual objectives often introduce stronger optimization conflicts with the original reconstruction losses, making training more unstable and difficult. This conflict can lead to severe rendering artifacts, as observed in the results of 3DGS and AbsGS in Figure~\ref{fig:intro}, Figure~\ref{fig:qualitative_results_wdr}, and Figure~\ref{fig:qualitative_results_lpips}. In contrast, our method mitigates these artifacts while preserving high-quality rendering details. As shown in the marked regions, GC-GS recovers fine-scale structures and lighting details more faithfully, including clearer reflectance details and more coherent structures of small objects such as the monitor and the sign.

\noindent\textbf{Efficiency Comparison.}
Table~\ref{tab:efficiency_results} provides a quantitative comparison of
rendering efficiency and training time. GC-GS achieves a clear advantage
in rendering speed by producing a more compact and effectively distributed
Gaussian representation. Although computing objective-specific gradients
and conflict-aware updates introduces additional optimization overhead,
our density-control strategy prevents excessive Gaussian growth and
reduces the subsequent rendering and optimization cost. Consequently,
GC-GS provides higher rendering efficiency while incurring moderate
training overhead.
\subsection{Ablation Study}

\noindent\textbf{Effectiveness of Geometry-Level Gradient Reconciliation.}
We first evaluate whether existing gradient manipulation methods can directly
benefit 3DGS optimization. As shown in the second and third rows of
Table~\ref{tab:ablation}, simply applying MGDA~\cite{mgda},  PCGrad~\cite{pcgrad} or CAGrad~\cite{cagrad}
to the baseline does not improve reconstruction quality and even degrades all
metrics. This suggests that directly modifying the global gradient is not well
suited for 3DGS: severe conflicts often occur only on a small subset of
Gaussians, while many primitives already receive compatible supervision. A
global gradient surgery may therefore make the overall update overly
conservative, suppress useful densification signals, or perturb otherwise
well-aligned Gaussian updates. In contrast, our Geometry-Level Gradient
Reconciliation (GLGR), shown in the fourth row, performs only mild position
adjustment on Gaussians with severe gradient conflict. By localizing the
intervention to conflicted geometry updates while preserving the original
updates for other primitives, GLGR improves all metrics over the baseline.

\noindent\textbf{Effectiveness of Conflict-Guided Density Control.}
Building on GLGR, we add Conflict-Guided Directional Densification (CGD),
corresponding to the fifth row. The further improvement demonstrates that
gradient conflict is not merely an optimization artifact, but also an effective
signal for identifying Gaussians whose local representation capacity is
insufficient. By encouraging these primitives to split or clone, CGD better
allocates Gaussians to difficult regions and improves reconstruction quality.
Finally, adding Conflict-Guided Opacity Suppression (CGOS) gives the full model
in the last row. The full model further improves the results, with a notable
PSNR gain of 0.07. This indicates that CGOS does more than reduce the number of
Gaussians: it helps identify primitives whose geometry remains unreliable near
convergence and suppresses their rendering contribution, leading to better scene
optimization.

\begin{table}[!h]
\centering
\caption{Ablation of various modules of ours and simple gradient surgery method. All
metrics are evaluated on the Mip-NeRF 360 dataset and averaged
across all scenes.
}
\label{tab:ablation}
\setlength{\tabcolsep}{4pt}
\begin{tabular}{lccc}
\toprule
Method & PSNR $\uparrow$ & SSIM $\uparrow$ & LPIPS $\downarrow$ \\
\midrule
Baseline &27.82  &0.827  &0.201  \\
Baseline + MGDA & 27.80  & 0.827  & 0.202  \\
Baseline + PCGrad & 27.61  & 0.812  & 0.209  \\
Baseline + CAGrad & 27.83  & 0.825  & 0.203  \\
\midrule
+ GLGR & 27.93 & 0.830  & 0.196  \\
+ GLGR + CGD & 28.05  & 0.836  & 0.183  \\
Ours full & 28.12  & 0.839 & 0.180  \\
\bottomrule
\end{tabular}
\end{table}

%% file: tables/perceptual_loss_results.tex
\newcommand{\best}[1]{\cellcolor{best}{#1}}
\newcommand{\second}[1]{\cellcolor{second}#1}
\begin{table*}[!t]
\centering
\small
\caption{Quantitative comparison on all three datasets for perceptual loss variants (with WD-R and LPIPS). We report PSNR$\uparrow$, SSIM$\uparrow$, LPIPS$\downarrow$, and DISTS$\downarrow$. The \colorbox{best}{best} and\colorbox{second}{second-best} results are highlighted.}
\setlength{\tabcolsep}{3pt}
\renewcommand{\arraystretch}{1.08}
\label{tab:quantitative_perceptual_loss_results}
\begin{tabular*}{\textwidth}{@{\extracolsep{\fill}}lcccccccccccc@{}}
\toprule
Method
& \multicolumn{4}{c}{Mip-NeRF 360} & \multicolumn{4}{c}{Tanks \& Temples} & \multicolumn{4}{c}{Deep Blending} \\
\cmidrule(lr){2-5} \cmidrule(lr){6-9} \cmidrule(lr){10-13}
& PSNR$\uparrow$ & SSIM$\uparrow$ & LPIPS$\downarrow$ & DISTS$\downarrow$ & PSNR$\uparrow$ & SSIM$\uparrow$ & LPIPS$\downarrow$ & DISTS$\downarrow$ & PSNR$\uparrow$ & SSIM$\uparrow$ & LPIPS$\downarrow$ & DISTS$\downarrow$ \\
\midrule
3DGS-WD-R & 27.04 & 0.791 & \second{0.174} & \second{0.041} & 23.08 & 0.822 & \second{0.138} & \second{0.039} & 29.31 & \second{0.891} & \best{0.193} & \second{0.067} \\
AbsGS-WD-R & 26.09 & 0.768 & 0.190 & 0.043 & 22.75 & 0.814 & 0.139 & 0.042 & 26.81 & 0.837 & 0.267 & 0.117 \\
Ours-WD-R & 27.41 & \second{0.812} & \best{0.160} & \best{0.037} & \second{23.43} & 0.824 & \best{0.137} & \best{0.036} & \best{29.85} & 0.890 & 0.204 & \best{0.060} \\
\midrule
3DGS-LPIPS & \second{27.46} & \second{0.812} & 0.176 & 0.045 & 23.42 & \second{0.835} & 0.157 & 0.048 & 29.18 & \best{0.892} & 0.203 & 0.078 \\
AbsGS-LPIPS & 26.00 & 0.767 & 0.215 & 0.051 & 23.27 & 0.834 & 0.158 & 0.045 & 24.80 & 0.798 & 0.327 & 0.170 \\
Ours-LPIPS & \best{27.58} & \best{0.814} & 0.175 & 0.044 & \best{23.75} & \best{0.838} & 0.169 & 0.045 & \second{29.47} & 0.885 & \second{0.201} & 0.071 \\
\bottomrule
\end{tabular*}%

\end{table*}

%% file: Figure/quanlitative_results_wdr.tex
\begin{figure*}[t]
    \centering
    \includegraphics[width=\textwidth]{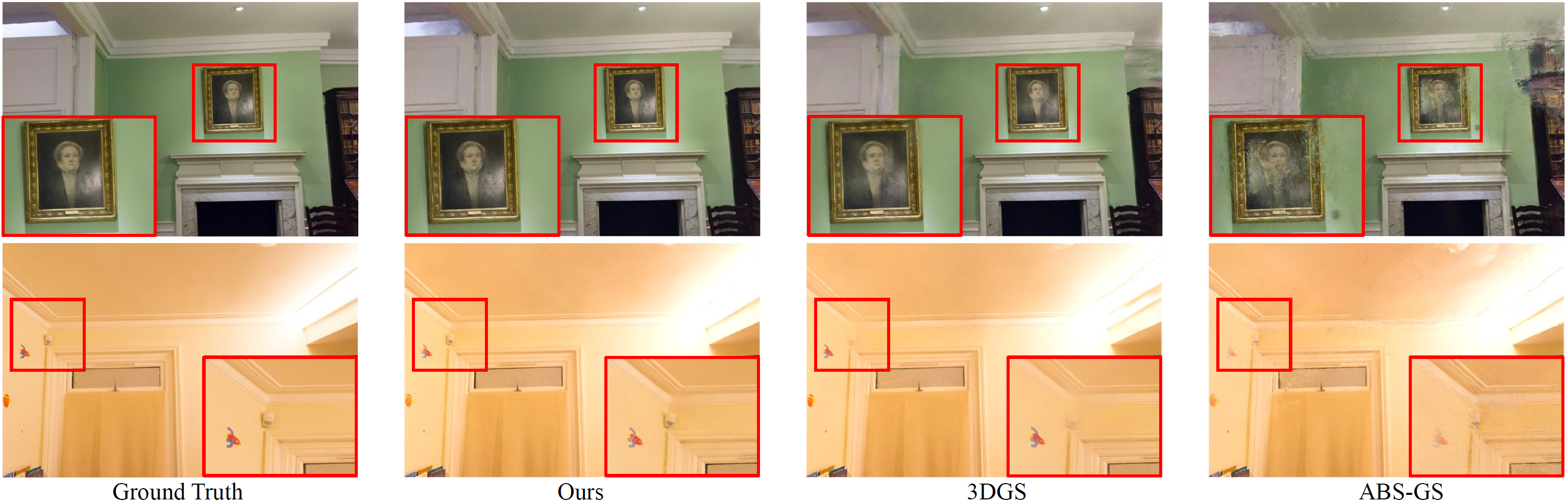}
    \caption{Qualitative comparisons under WD-R~\cite{dropin} loss.
GC-GS suppresses rendering artifacts and preserves finer visual details
compared with other methods.}
    \label{fig:qualitative_results_wdr}
\end{figure*}

%% file: Figure/quanlitative_results_lpips.tex
\begin{figure*}[t]
    \centering
    \includegraphics[width=\textwidth]{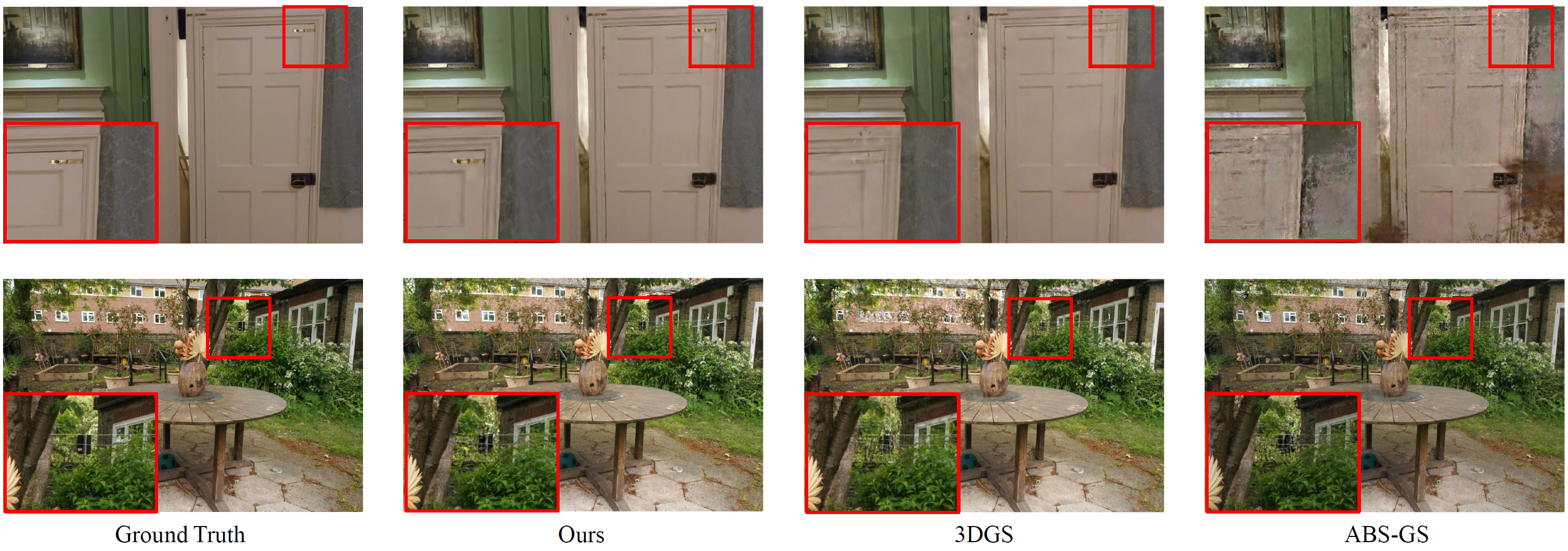}
    \caption{Qualitative comparisons under LPIPS~\cite{dropin} loss.
GC-GS suppresses rendering artifacts and preserves finer visual details
compared with other methods.
}
    \label{fig:qualitative_results_lpips}
\end{figure*}

%% file: sec/5_conclusion.tex
\section{Conclusion}
In this work, we presented GC-GS, a 3DGS framework that explicitly
exploits conflicts between reconstruction objectives. We formulate 3DGS
training from a multi-objective optimization perspective and selectively reconcile the gradients of highly conflicting primitives, preventing
opposing objectives from dominating their position updates. We further
use persistent position-gradient conflict as a signal for adaptive density
control. During early training, conflict-guided splitting and cloning
increase the local representation capacity of under-reconstructed
regions. During late training, Gaussians exhibiting persistent conflict
and high rendering error are progressively suppressed through opacity
attenuation, reducing their unreliable rendering contributions.
Experiments demonstrate that GC-GS achieves superior rendering quality
with fewer Gaussians, consistently improves performance across
conventional and perceptual training objectives, and incurs moderate training
overhead. Future work may explore the connection between pixel-level loss
disagreement and primitive-level conflict, and extend GC-GS to
compression-oriented and dynamic 3DGS settings with additional objectives
such as bitrate and temporal consistency.